\documentclass{article}

\usepackage{microtype}
\usepackage{graphicx}
\usepackage{subcaption}
\usepackage{booktabs} 
\usepackage[table]{xcolor} 

\usepackage{hyperref}

\usepackage[preprint]{icml2027}
\usepackage{amssymb}
\usepackage{mathtools}
\usepackage{amsthm}
\usepackage[most]{tcolorbox}
\newtcolorbox{mybox}[1][]{%
  colback=gray!5!white,   
  colframe=gray!75!black, 
  boxrule=0.8pt,
  arc=2mm,
  left=4mm, right=4mm, top=2mm, bottom=2mm,
  enhanced,
  fonttitle=\bfseries\large, 
  fontupper=\large,        
  title={#1}
}

\usepackage{amsmath}
\usepackage{amssymb}
\usepackage{mathtools}
\usepackage{amsthm}

\usepackage[capitalize,noabbrev]{cleveref}

\theoremstyle{plain}

\theoremstyle{definition}

\theoremstyle{remark}

\usepackage[disable,textsize=tiny]{todonotes}

\usepackage{url}
\usepackage{capt-of}

\usepackage[utf8]{inputenc}
\usepackage[T1]{fontenc}
\usepackage{amsfonts}
\usepackage{nicefrac}
\usepackage{multirow}
\usepackage{pifont}
\usepackage{tikz}
\usetikzlibrary{arrows.meta}

\usepackage{algorithm}
\usepackage{algpseudocodex}
\usepackage{amsmath}
\usepackage{bbm}
\usepackage{textcomp}
\usepackage{stfloats}
\usepackage{mathabx}
\usepackage{enumitem}
\setlist{leftmargin=5mm}
\usepackage{wrapfig}
\usepackage{makecell}
\usepackage{lipsum}
\usepackage{xcolor,xspace}
\usepackage{float}
\usepackage{arydshln}

\definecolor{correctgreen}{RGB}{0, 153, 0}
\definecolor{wrongred}{RGB}{204, 0, 0}

\definecolor{dpcol}{RGB}{217, 196, 230}
\definecolor{sgcol}{RGB}{198, 222, 242}
\definecolor{nbcol}{RGB}{207, 232, 200}

\tcbuselibrary{breakable} 

\newtcolorbox{ReasoningBox}[1][]{
    enhanced,
    breakable, 
    colback=white,
    colframe=black,
    coltitle=white,
    fonttitle=\bfseries,
    title=Reasoning Process Error,
    sharp corners,
    boxrule=1pt,
    attach boxed title to top left={xshift=0mm, yshift=0mm},
    boxed title style={
        sharp corners, 
        colback=black, 
        colframe=black, 
        boxrule=1pt,
        right=2mm, left=2mm
    },
    top=1.5em,
    #1
}

\makeatletter
\newcommand{\DrawLine}{%
  \begin{tikzpicture}
  \path[use as bounding box] (0,0) -- (\linewidth,0);
  \draw[color=black,dashed,dash phase=2pt]
        (0-\kvtcb@leftlower-\kvtcb@boxsep,0)--
        (\linewidth+\kvtcb@rightlower+\kvtcb@boxsep,0);
  \end{tikzpicture}%
  }
\definecolor{violet}{RGB}{138, 43, 226}

\makeatother

\definecolor{citecol}{HTML}{2DDC0E}
\definecolor{tableofcontent}{HTML}{E63E15}
\definecolor{urlcol}{HTML}{2470D8}
\definecolor{myorange}{RGB}{2, 142, 2}
\hypersetup{
    colorlinks=true,
    linkcolor=red,
    filecolor=magenta,      
    urlcolor=cyan,
    citecolor=myorange,
}

\tcbset{
  colback=gray!5!white,    %
  colframe=black,          %
  boxrule=0.5pt,           %
  arc=2pt,                 %
  left=6pt, right=6pt, top=4pt, bottom=4pt,
  fonttitle=\bfseries,
  coltitle=black
}
\usepackage[capitalize,noabbrev]{cleveref}

\theoremstyle{plain}

\theoremstyle{definition}

\theoremstyle{remark}

\newcommand\ours{\textsc{EpiCompass}\xspace}
\usepackage[textsize=tiny]{todonotes}

\icmltitlerunning{Steering Automated Scientific Discovery via Audited Epistemic Exploration}

\begin{document}

\twocolumn[
\icmltitle{Steering Automated Scientific Discovery via Audited Epistemic Exploration}



  \icmlsetsymbol{equal}{*}

  \begin{icmlauthorlist}
\icmlauthor{Fangxu Yu}{yyy}
\icmlauthor{Tao Feng}{xxx}
\icmlauthor{Xuying Ning}{xxx}
\icmlauthor{Haolin Yang}{aaa}
\icmlauthor{Jiaxuan You}{xxx}
\icmlauthor{Tianyi Zhou}{aaa}
  \end{icmlauthorlist}

  \icmlaffiliation{yyy}{University of Maryland, College Park}
\icmlaffiliation{aaa}{Mohamed bin Zayed University of Artificial Intelligence}
\icmlaffiliation{xxx}{University of Illinois at Urbana–Champaign}

\icmlcorrespondingauthor{Tianyi Zhou}{tianyi.zhou@mbzuai.ac.ae}
\icmlsetsymbol{dag}{\dag}

  \icmlkeywords{Machine Learning, ICML}

  \vskip 0.3in
]



\printAffiliationsAndNotice{}  

\begin{abstract}
Scientific discovery requires revisiting the hypotheses in light of new evidence and identifying promising directions beyond those already explored. However, research agents often make claims that are not supported by their observations, explore a narrow hypothesis space, and retain beliefs despite recognizing conflicting evidence. These failures motivate the search for an explicit epistemic state that captures both beliefs about existing hypotheses and directions for further exploration. To this end, we propose Epistemic Compass (\ours), a plug-and-play scaffold that maintains this state through an epistemic tree linking hypotheses, evidence, beliefs, and revisions. As an investigation progresses, \ours{} reassesses the hypotheses explored against accumulated evidence and uses unresolved conflicts and unexplored directions to guide hypothesis generation. It selects candidates by anticipated epistemic gain, accounting for both refinement of existing hypotheses and expansion of the hypothesis space, and verifies the final conclusion against the accumulated evidence. Experimental results across three scientific discovery benchmarks and three LLM backbones demonstrate that \ours consistently outperforms baselines, with further analyses showing reductions in errors of unsupported claims and failures to revise beliefs. Our code is available at \href{https://github.com/tianyi-lab/EpiCompass}{https://github.com/tianyi-lab/EpiCompass}.
\end{abstract}

\section{Introduction}\label{sec:intro}

Scientific discovery differs from conventional problem solving in how its questions and objectives evolve. Whereas problem solving typically begins with a well-defined question and a specified goal, discovery continually revises what to ask as evidence accumulates. Progress therefore involves advancing the epistemic frontier. Identifying a previously unrecognized possibility turns an \emph{unknown unknown} into a \emph{known unknown}; testing it can establish a provisionally accepted explanation, or \emph{known known}. An unexpected observation can, in turn, reopen a question previously considered settled. Discovery thus requires more than verifying candidate answers. It also requires identifying questions worth asking and reconsidering assumptions that no longer fit the evidence. Supporting this process calls for an explicit, continually updated epistemic state with two complementary components: a \emph{belief state} that tracks confidence in proposed hypotheses and their supporting or conflicting evidence, and an \emph{exploration state} that tracks unresolved gaps, unexplained observations, and promising directions for generating new hypotheses.

Recent research agents have made substantial progress in organizing hypotheses and evidence and maintaining explicit belief states. Hypothesis registries and trees record candidate explanations and their relationships to evidence, with some also tracking beliefs or hypothesis status~\citep{takahara2026auditable,jin2026arbor,wu2026bayesevolve,chen2026statefuldiscovery}. Related approaches maintain belief representations during long-horizon tasks~\citep{luo2026beyond,singh2026brace} or guide experimentation through assessments of how observations may change beliefs over candidate explanations~\citep{agarwal2026autodiscovery,agarwal2026evidence,murphy2026model,yuh2026bayesian}. These advances raise a further question: how should evolving beliefs guide exploration beyond the hypotheses already considered? New observations can alter the credibility of existing explanations, but they can also reveal previously unrecognized gaps and suggest new directions for investigation. However, making these gaps explicit and using them to guide hypothesis generation remains challenging. When unresolved questions remain implicit in the interaction history, agents must reconstruct them when deciding what to investigate next. Reports of neglected evidence and limited revision after refutation further motivate examining how agents connect accumulated evidence to subsequent investigation~\citep{rios2026ai,bisht2026agentic}.

\begin{table*}[t]
\centering
\caption{\textbf{Design comparison with representative discovery agent methods.} Candidate exploration denotes sampling and comparing multiple hypotheses at a proposal step; belief tracking denotes explicit confidence over explored hypotheses; exploration state denotes recorded unresolved gaps and prospective directions; epistemic-gain selection denotes assessment of both belief refinement and exploration potential; evidence-grounded verification denotes a dedicated check before final submission. Marks refer to explicit framework components.}
\label{tab:comparison}
\resizebox{\textwidth}{!}{%
\begin{tabular}{lccccc}
\toprule
 & Candidate Exploration & Belief Tracking & Exploration State & Epistemic-Gain Selection & Evidence-Grounded Verification \\
\midrule
ReAct~\citep{yao2022react} & \textcolor{wrongred}{\ding{55}} & \textcolor{wrongred}{\ding{55}} & \textcolor{wrongred}{\ding{55}} & \textcolor{wrongred}{\ding{55}} & \textcolor{wrongred}{\ding{55}} \\
Hypothesis Search~\citep{wang2024hypothesis} & \textcolor{correctgreen}{\ding{51}} & \textcolor{wrongred}{\ding{55}} & \textcolor{wrongred}{\ding{55}} & \textcolor{wrongred}{\ding{55}} & \textcolor{wrongred}{\ding{55}} \\
Popper~\citep{huang2025automated} & \textcolor{wrongred}{\ding{55}} & \textcolor{wrongred}{\ding{55}} & \textcolor{wrongred}{\ding{55}} & \textcolor{wrongred}{\ding{55}} & \textcolor{correctgreen}{\ding{51}} \\
HEP~\citep{takahara2026auditable} & \textcolor{wrongred}{\ding{55}} & \textcolor{correctgreen}{\ding{51}} & \textcolor{wrongred}{\ding{55}} & \textcolor{wrongred}{\ding{55}} & \textcolor{wrongred}{\ding{55}} \\
Arbor~\citep{jin2026arbor} & \textcolor{correctgreen}{\ding{51}} & \textcolor{wrongred}{\ding{55}} & \textcolor{wrongred}{\ding{55}} & \textcolor{wrongred}{\ding{55}} & \textcolor{wrongred}{\ding{55}} \\
\ours{} (Ours) & \textcolor{correctgreen}{\ding{51}} & \textcolor{correctgreen}{\ding{51}} & \textcolor{correctgreen}{\ding{51}} & \textcolor{correctgreen}{\ding{51}} & \textcolor{correctgreen}{\ding{51}} \\
\bottomrule
\end{tabular}%
}
\vspace{-10pt}
\end{table*}

To examine this challenge in practice, our initial error analysis suggests that the main bottleneck lies beyond implementing experiments (\S\ref{sec:failure-modes}). The key shortfall lies in how agents interpret experimental evidence, revise their beliefs, and decide which hypotheses to explore next. Agents make claims unsupported by their own observations, remain confined to narrow hypothesis spaces, and fail to revise beliefs even after recognizing conflicting evidence. These findings suggest that maintaining a record of existing hypotheses is only part of the challenge, as agents must also use their evolving epistemic state to identify unresolved questions and guide further exploration.

Motivated by this, we propose Epistemic Compass (\ours), a plug-and-play scaffold that maintains an explicit epistemic state to guide scientific discovery. It represents the investigation as an epistemic tree, where nodes record hypotheses, beliefs, and their relationships to evidence, while edges trace revisions and the conflicts they resolve or introduce. After each experiment or hypothesis proposal, \ours{} updates beliefs against the accumulated evidence and identifies exploration directions from unresolved conflicts and unexplored branches. This evolving state guides hypothesis generation through \textit{epistemic gain}, which considers both progress on existing hypotheses and opportunities to expand or revise the hypothesis space, complementing criteria based on belief change or uncertainty over current candidates. The agent generates multiple candidates conditioned on the tree and selects one by weighing evidential support and the novelty of its proposed direction. Before final submission, it reviews the complete tree to decide whether to revise its explanation, experiment further, or confirm its conclusion. Table~\ref{tab:comparison} highlights the key differences between \ours{} and representative discovery agent methods in hypothesis exploration, epistemic state tracking, and evidence-grounded verification.

We evaluate \ours{} on DiscoverPhysics~\citep{wiemann2026discoverphysics}, ActiveSciBench-Chem~\citep{kabra2026llm}, and NewtonBench~\citep{zheng2026newtonbench} with three LLM backbones. Under the same experimental budgets, \ours{} improves the primary discovery metric across benchmarks and backbones. Compared with the strongest baseline in each setting, it reduces normalized trajectory error on DiscoverPhysics by up to 94.1\% and achieves relative accuracy gains of up to 21.0\% on ActiveSciBench-Chem and 19.9\% on NewtonBench. 
Further analyses show that \ours{} reduces unsupported claims and failures to revise beliefs despite conflicting evidence, while encouraging further investigation and hypothesis revision before submission. Ablation studies support the complementary contributions of hypothesis selection, belief updates, and final self-verification. 

\section{Methodology}\label{sec:method}

In this section, We characterize common discovery failures (\S\ref{sec:failure-modes}), then formulate the task (\S\ref{sec:prob_form}) and introduce \ours{} (\S\ref{sec:ours}). 

\begin{figure}[t]
\includegraphics[width=1.0\columnwidth]{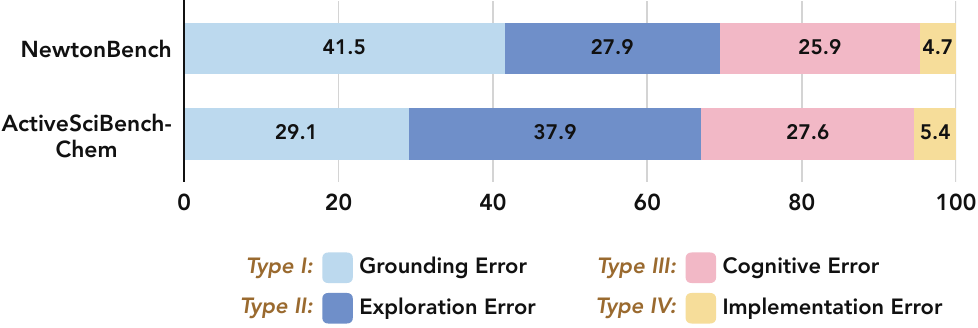}
\caption{Distribution (\%) of four error types among categorized failures of Gemini-3.1-Flash-Lite on NewtonBench and ActiveSciBench-Chem. Each benchmark is normalized to 100\%; these proportions differ from the rates over all trajectories in Figure~\ref{fig:error}.}
\label{fig:failure-modes}
\vspace{-10pt}
\end{figure}

\subsection{Error Analysis of Automated Scientific Discovery}\label{sec:failure-modes}
We first analyze failed discovery trajectories of the ReAct agent with Gemini-3.1-Flash-Lite on NewtonBench and ActiveSciBench-Chem to characterize obstacles to successful discovery. Figure~\ref{fig:failure-modes} reports the relative frequencies of the four error categories, normalized to sum to 100\% within each benchmark.
We identify four dominant error types: (1) \textbf{Grounding Error}: Claims unsupported by the agent's own data, such as misreading or fabricating experimental outcomes; (2) \textbf{Exploration Error}: Exploration confined to a narrow hypothesis space or failing to pursue relevant directions; (3) \textbf{Cognitive Error}: Failure to revise belief in a hypothesis despite recognizing conflicting evidence; (4) \textbf{Implementation Error}: Incorrect implementation of a sound idea, such as runtime errors, mistranscribed formulas, or invalid submission formats. As shown in Figure~\ref{fig:failure-modes}, grounding errors dominate on NewtonBench, whereas exploration errors are most prevalent on ActiveSciBench-Chem. Cognitive errors account for roughly a quarter of failures on both benchmarks, indicating a recurring difficulty in revising beliefs despite recognized conflicts with evidence. These findings motivate explicitly tracking hypotheses, evidence, and evolving beliefs, and using this epistemic state to guide subsequent investigation.

\subsection{Problem Formulation}\label{sec:prob_form}
We frame automated discovery as active experimentation under a fixed budget. Let $\mathcal{H}$ be a space of executable hypotheses, each defining a predictive mapping $h: \mathcal{X} \to \mathcal{Y}$ from experiments to outcomes, and let $f^\star$ be the hidden mechanism to be recovered. Running an experiment $x \in \mathcal{X}$ yields an outcome $y \sim P_{\text{Env}}(\cdot \mid x, f^\star)$ returned by the environment. For example, in NewtonBench, $f^\star$ may be a counterfactually shifted law of gravitation; an experiment $x$ specifies the masses and distance of two bodies, and its outcome $y$ is the measured force. A discovery agent is an LLM-based policy $\pi$. At each step $t$, it takes an action $a_t \sim \pi(\cdot \mid \tau_{t-1})$ given its trajectory $\tau_{t-1} = (a_1, o_1, \dots, a_{t-1}, o_{t-1})$. During investigation, the agent performs one of two actions: \textit{hypothesis proposal}, which introduces a new executable hypothesis $h_t \in \mathcal{H}$ or revises an existing one; or \textit{experimental testing}, which runs an experiment $x_t$ and receives an observation containing its outcome $y_t$. Experimental testing can assess existing hypotheses or explore unresolved questions. It may also request final submission of a hypothesis. The episode ends with a final hypothesis $\hat{h}$ under the benchmark's termination rules and experiment budget $B$. The objective is to minimize the expected discovery error $\mathbb{E}\big[\delta(\hat{h}, f^\star)\big]$, where $\delta$ is a domain-dependent error measure and the expectation is taken over the trajectories induced by the agent and the environment. Since $f^\star$ remains hidden throughout the process, the agent must evaluate its hypotheses using the experiments it has conducted.

\subsection{Overview}\label{sec:ours}
\begin{figure*}[t]
\includegraphics[width=1.0\textwidth]{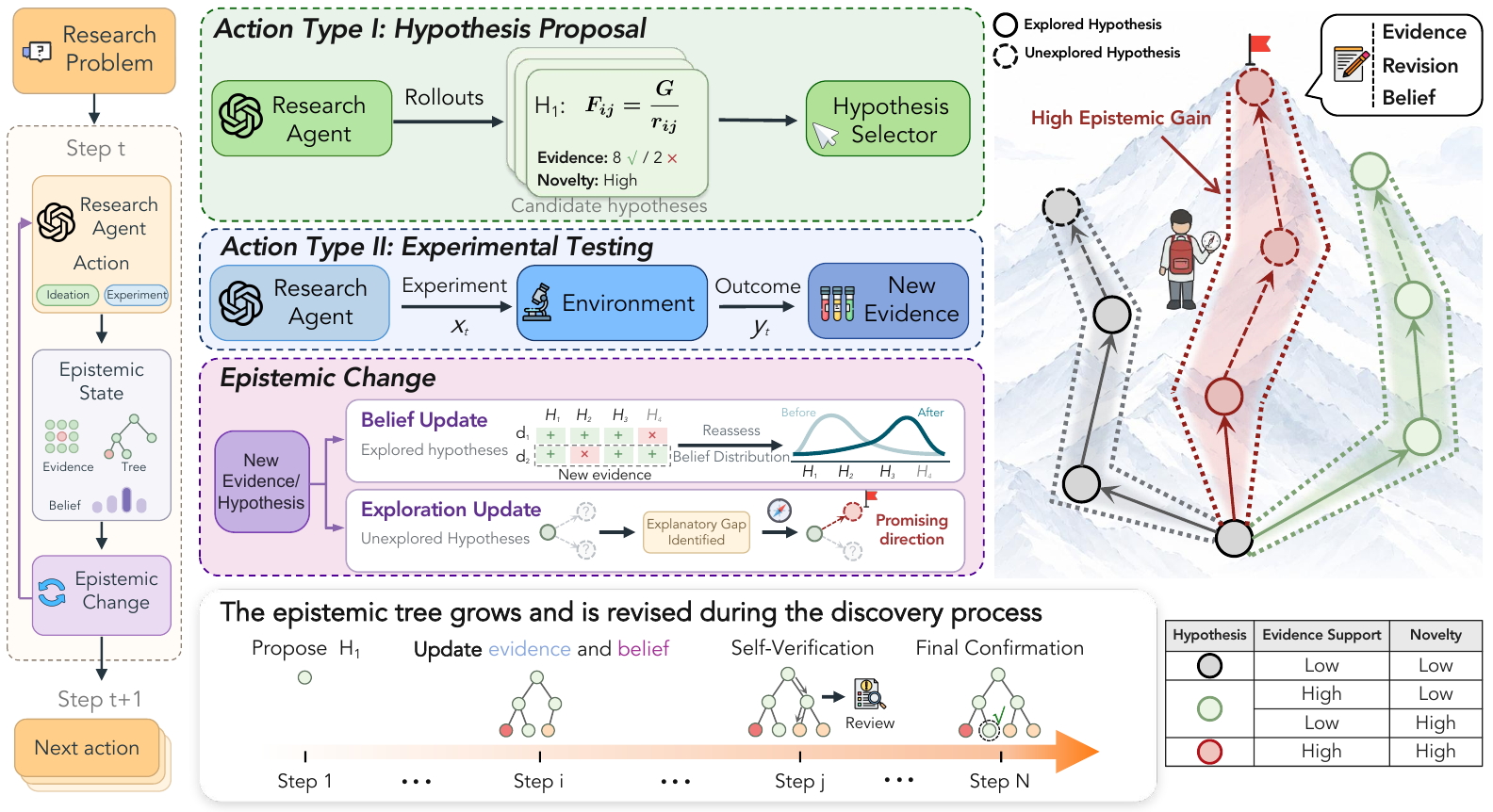}
\caption{Overview of \ours{}. Hypothesis proposal and experimental testing change the agent's epistemic state through two complementary pathways: \textit{belief update} reassesses explored hypotheses against accumulated evidence, while \textit{exploration update} identifies and refines promising directions in the unexplored hypothesis space. The updated state guides subsequent proposals and experiments. Hypothesis selection considers evidence support and novelty as indicators of anticipated epistemic gain, and final self-verification checks the proposed explanation against the complete epistemic tree.}
\label{fig:overview}
\vspace{-10pt}
\end{figure*}

Scientific discovery requires updating both the credibility of existing explanations and the directions worth exploring next. As shown in Figure~\ref{fig:overview}, \ours{} connects \textit{hypothesis proposal} and \textit{experimental testing} through an evolving epistemic state represented by an epistemic tree (\S\ref{sec:tree}). New evidence or a proposed hypothesis triggers two complementary updates: \textit{belief update} reassesses explored hypotheses, while \textit{exploration update} identifies and refines unexplored directions suggested by explanatory gaps (\S\ref{sec:belief}). During hypothesis proposal, \ours{} samples candidate hypotheses and selects one according to its anticipated epistemic gain, considering both evidential support and the value of the direction it introduces (\S\ref{sec:sampling}). Experimental testing assesses existing explanations and investigates unresolved questions. Before final confirmation, epistemically grounded self-verification revisits the complete tree (\S\ref{sec:verify}). Algorithm~\ref{alg:ours} in Appendix~\ref{app:algorithm} summarizes the procedure.

\subsection{Epistemic State Representation}\label{sec:tree}
We represent the agent's epistemic state as $\mathcal{E}_t=(\mathcal{D}_t,H_t,b_t,U_t)$, where $\mathcal{D}_t\subseteq\mathcal{X}\times\mathcal{Y}$ contains observed experiment--outcome pairs, $H_t\subseteq\mathcal{H}$ contains hypotheses retained in the investigation, and $b_t\in\Delta(H_t)$ represents normalized beliefs over these candidates when $H_t$ is nonempty. An \textit{explored} hypothesis has been retained in the investigation and assessed against available evidence. The component $U_t$ captures recognized directions toward unexplored hypotheses, including explanatory gaps and possible mechanisms suggested by observations or revisions. It represents where further exploration may be informative, without enumerating $\mathcal{H}\setminus H_t$ or assigning beliefs to unseen hypotheses.

\noindent\textbf{Hypothesis--Evidence Relationships.}
For each hypothesis $h \in H_t$ and experiment $d \in \mathcal{D}_t$, we represent their relationship by
\begin{equation}
v(h,d) \in \{\text{support}, \text{conflict}, \text{neutral}\},
\label{eq:label}
\end{equation}
where \textit{support} and \textit{conflict} indicate evidence for or against the hypothesis, respectively, and \textit{neutral} indicates that the observation is orthogonal to the hypothesis. We collect these labels into $V_t = \{v(h,d)\}_{h \in H_t,\, d \in \mathcal{D}_t}$. Together, $V_t$ and $b_t$ describe the evidential status of the candidates, with the \textit{main hypothesis} chosen as $m_t \in \arg\max_{h \in H_t} b_t(h)$. 

\noindent\textbf{Epistemic Tree.}
\ours{} organizes hypotheses into an epistemic tree $\mathcal{T}_t$. Each node stores a hypothesis's content, belief $b_t(h)$, and evidence labels. Each edge records the relation between a child and its parent, including which conflicts the child resolves and which previously supported observations it no longer explains. A new hypothesis is added as a child of an existing hypothesis node if it builds on that hypothesis. Conflicting observations and unresolved branches ground the directions in $U_t$. 

\subsection{Belief and Exploration Updates}\label{sec:belief}
New evidence can change the credibility of an explored hypothesis while revealing directions that the agent had not previously recognized as worth investigating. A newly proposed hypothesis can likewise reorganize the interpretation of existing evidence and expose further explanatory gaps. We therefore update the epistemic state through two complementary pathways:
\begin{equation}
\underbrace{b_{t-1}\longrightarrow b_t}_{\text{belief update}},
\qquad
\underbrace{U_{t-1}\longrightarrow U_t}_{\text{exploration update}}.
\label{eq:state_update}
\end{equation}
Both updates are grounded in the same accumulated hypothesis--evidence relationships. They determine not only which explanation the agent currently favors, but also which unresolved directions should inform its subsequent hypothesis proposals and experimental tests.

\noindent\textbf{Hypothesis Check.}
When the agent conducts an experiment $x_t$ and observes $y_t$, \ours{} adds the observed pair to the evidence record, $\mathcal{D}_t = \mathcal{D}_{t-1} \cup \{(x_t,y_t)\}$. It then replays the hypotheses on the accumulated experiments to update the labels in $V_t$. Similarly, each new hypothesis is evaluated against all accumulated evidence.

\noindent\textbf{Belief Update over Explored Hypotheses.}
The evidence labels describe how individual observations bear on each hypothesis. To integrate these assessments, the agent reads the hypotheses and their evidence labels and updates a belief distribution:
\begin{equation}
b_t = \textsc{Belief}\big(H_t,V_t\big),
\label{eq:belief}
\end{equation}
where $b_t(h)$ represents the credence assigned to $h$ among the current candidates and $\sum_{h \in H_t} b_t(h)=1$, which are LLM-assigned relative credences. The update aims to discriminate among competing explanations as evidence accumulates. Following the belief update, $m_t$ is revised if another hypothesis attains the highest belief score.

\noindent\textbf{Exploration Update toward Unexplored Hypotheses.}
The updated evidence relationships also refine $U_t$, identifying where further exploration may be informative. Observations that conflict with current hypotheses expose explanatory gaps, while revisions that resolve or introduce conflicts clarify which assumptions warrant reconsideration. Further evidence can strengthen a direction's relevance, narrow its scope, or resolve the gap that motivated it. Evidence labels and revision records convey these directions through the state summary, guiding subsequent hypothesis proposals and experimental tests. Thus, previously unrecognized limitations become explicit questions, providing a more focused understanding of the unexplored hypothesis space.

\paragraph{Epistemic Feedback.}
We provide the agent with a structured summary $s_t$ of the current epistemic tree, including leading hypotheses, their beliefs, supporting or conflicting evidence, and unresolved directions for exploration. The summary highlights changes in the main hypothesis and guides subsequent actions together with the interaction history. 

\subsection{Hypothesis Exploration with Epistemic Gain}
\label{sec:sampling}
\noindent\textbf{Epistemic Gain.}
We describe epistemic gain as progress in the agent's epistemic state along two complementary pathways:
\begin{equation}
\begin{aligned}
G_{\mathrm{epi}}(\mathcal{E}_{t-1},\mathcal{E}_t)
={}&G_{\mathrm{belief}}(b_{t-1},b_t)\\
&+G_{\mathrm{exploration}}(U_{t-1},U_t).
\end{aligned}
\label{eq:epistemic_gain}
\end{equation}
The \textit{belief} component captures evidence-grounded reassessment of explored hypotheses, including increasing or decreasing the belief assigned to a hypothesis when warranted by evidence. Belief change alone is not necessarily progress. The \textit{exploration} component captures improved directionality in the unexplored hypothesis space: recognizing explanatory gaps, narrowing candidate mechanisms, or identifying informative observations. 

\noindent\textbf{Candidate Generation and Selection.}
Agents may remain committed to an initial hypothesis despite evidence suggesting alternative explanations~\citep{kirgis2026can, fei2026agents}. To support revision and exploration, \ours{} samples $K$ candidates conditioned on the interaction history and latest epistemic feedback:
\begin{equation}
h_t^{(1)},\dots,h_t^{(K)}\sim\pi(\cdot\mid\tau_{t-1},s_{t-1}).
\label{eq:sample}
\end{equation}
The LLM assesses each candidate's \textit{anticipated epistemic gain} through its evidence labels on $\mathcal{D}_{t-1}$, content, and relation to existing hypotheses. These criteria provide a qualitative assessment of potential belief refinement and exploration value. Then, the LLM selects
\begin{equation}
h_t=\textsc{Select}\big(\{h_t^{(k)}\}_{k=1}^{K};\mathcal{E}_{t-1}\big).
\label{eq:select}
\end{equation}
The selected hypothesis is added to $H_t$ and the epistemic tree, and is evaluated against accumulated evidence to update both beliefs and exploration directions.
\subsection{Epistemically Grounded Self-Verification}\label{sec:verify}
Although epistemic state tracking supports hypothesis revision during investigation, an agent may still commit to a final result before sufficient evidence distinguishes it from plausible alternatives. To alleviate this issue, \ours{} presents the complete epistemic tree, including previously reported evidence and revisions, to the agent when it first submits a hypothesis $\hat{h}$. The agent then checks its proposed hypothesis against accumulated observations and competing hypotheses, and revisits the unresolved questions exposed by the tree. It may revise its answer, run further experiments within the remaining budget, or confirm the answer unchanged. The next submission is final. This step connects the decision to conclude with both the evidence for the proposed explanation and the remaining reasons to continue investigation.

\section{Experiments}\label{sec:exp}

\subsection{Setup}\label{sec:setup}
In our experiments, we use three interactive discovery benchmarks that span physical law discovery and enzyme kinetics. In all of them, the agent interacts with an experimental environment whose mechanism is hidden, designs its own experiments, and submits an executable hypothesis that is scored against the hidden mechanism. DiscoverPhysics~\citep{wiemann2026discoverphysics} asks the agent to discover the laws of motion of simulated worlds whose physics deviates from our own, by observing noisy particle trajectories and submitting both an executable law and a natural-language explanation. ActiveSciBench-Chem~\citep{kabra2026llm} asks the agent to discover the initial-rate law of an enzyme-catalyzed reaction as a function of the substrate, inhibitor, co-substrate, product, and enzyme concentrations, the temperature, and the pH. NewtonBench~\citep{zheng2026newtonbench} asks the agent to discover counterfactually shifted physical laws across 12 domains, embedded in vanilla equations, simple systems, and complex systems of increasing difficulty.

Experiments are conducted with three LLM backbones, and within each benchmark--backbone setting, all methods share the same task interface and experimental limits. The baselines are adapted to this common interaction loop. For each backbone, we consider three baselines:
(i) \textit{ReAct}~\citep{yao2022react} interleaves reasoning and actions in a simple loop, which keeps the comparison focused on how the agent handles its own evidence;
(ii) \textit{Hypothesis Search}~\citep{wang2024hypothesis} samples four candidate hypotheses whenever the agent proposes a hypothesis, and the same LLM selects the most promising one given the trajectory so far;
(iii) \textit{Popper}~\citep{huang2025automated} designs and runs falsification experiments against the agent's current hypothesis before submission, and returns to the agent whether each test passes or fails.

\subsection{Implementation Details}\label{sec:impl}

We use Gemini-3.1-Flash-Lite, Gemini-3.7-Flash, and Muse-Glimmer-30B as the backbones. \ours{} performs candidate selection and belief updates using the same backbone, samples $K=4$ candidate hypotheses at each proposal step, and limits routine epistemic feedback to eight presentations per episode.
For evaluation, we follow the official evaluation code of each benchmark and report the normalized trajectory MSE (nMSE), the explanation score (Expl.), and pass@$k$ on DiscoverPhysics, the exact accuracy (Acc.) and RMSLE on ActiveSciBench-Chem, and the symbolic accuracy (SA) and RMSLE on NewtonBench (see Appendix~\ref{app:benchmarks} for more details).

\subsection{Main Results}\label{sec:main}
Tables~\ref{tab:main} and~\ref{tab:newtonbench} present the performance of all methods on the three benchmarks with three LLM backbones. The best result for each metric within each backbone is shown in bold. We highlight two observations:

(i) \textbf{\ours{} consistently improves discovery performance across benchmarks and backbones.} Compared with ReAct, \ours{} increases symbolic accuracy on NewtonBench by 14.3 and 15.3 percentage points with Gemini-3.1-Flash-Lite and Gemini-3.7-Flash, corresponding to relative improvements of 30.7\% and 40.6\%, respectively. With these two backbones, it also reduces normalized trajectory error on DiscoverPhysics by factors of 17.9 and 10.1. On ActiveSciBench-Chem, \ours{} raises exact accuracy from 69.1\% to 85.2\% with Muse-Glimmer-30B. Overall, \ours{} improves the primary discovery metric for every benchmark--backbone combination. These gains span symbolic law recovery, prediction of physical trajectories, and identification of enzyme kinetics, supporting the applicability of epistemic state tracking across the evaluated settings.

(ii) \textbf{\ours{} generalizes across scientific domains.} \ours{} consistently improves discovery performance in both physics and chemistry across all three backbones. While the baselines also achieve gains on physics benchmarks, their improvements do not consistently extend to ActiveSciBench-Chem. Notably, \ours{} raises exact accuracy on ActiveSciBench-Chem from 69.1\% to 85.2\% with Muse-Glimmer-30B. Its gains therefore extend beyond recovering physical laws to identifying chemical reaction mechanisms, despite differences in experimental variables and hypothesis spaces. These results demonstrate the generality of \ours across distinct scientific domains.

\begin{table*}[t]
\centering
\caption{Experimental results of \ours{} and baseline methods on DiscoverPhysics and ActiveSciBench-Chem across three backbone LLMs. Explanation score ($\uparrow$), normalized trajectory MSE ($\downarrow$), and pass@$k$ ($\uparrow$) are reported on DiscoverPhysics; exact accuracy (Acc.~$\uparrow$) and RMSLE ($\downarrow$) are reported on ActiveSciBench-Chem.}
\label{tab:main}
\resizebox{0.9\textwidth}{!}{%
\begin{tabular}{l|ccccc|cc}
\toprule
\multirow{2}{*}{\textbf{Method}} & \multicolumn{5}{c|}{\cellcolor{dpcol}\textbf{DiscoverPhysics}} & \multicolumn{2}{c}{\cellcolor{sgcol}\textbf{ActiveSciBench-Chem}} \\
 & Expl. $\uparrow$ & nMSE $\downarrow$ & pass@1 (\%) $\uparrow$ & pass@2 (\%) $\uparrow$ & pass@3 (\%) $\uparrow$ & Acc. (\%) $\uparrow$ & RMSLE $\downarrow$ \\
\midrule
\rowcolor{gray!20} \multicolumn{8}{c}{\textit{Gemini-3.1-Flash-Lite}} \\
\midrule
ReAct & 0.300 & 25.46 & 0.0 & 0.0 & 0.0 & 64.2 & 0.023 \\
Hypothesis Search & \textbf{0.385} & 34.69 & 0.0 & 0.0 & 0.0 & 61.7 & 0.021 \\
Popper & 0.358 & 17.26 & 0.0 & 0.0 & 0.0 & 75.3 & 0.017 \\
\hdashline\noalign{\vskip 2pt}
\ours{} & 0.379 & \textbf{1.42} & \textbf{3.0} & \textbf{6.1} & \textbf{9.1} & \textbf{79.0} & \textbf{0.015} \\
\midrule
\rowcolor{gray!20} \multicolumn{8}{c}{\textit{Gemini-3.7-Flash}} \\
\midrule
ReAct & \textbf{0.539} & 0.121 & 6.1 & 12.1 & 18.2 & 87.7 & \textbf{0.006} \\
Hypothesis Search & 0.524 & 0.074 & 6.1 & 12.1 & 18.2 & 80.2 & 0.009 \\
Popper & 0.503 & 0.120 & \textbf{15.2} & 18.2 & 18.2 & 70.4 & 0.012 \\
\hdashline\noalign{\vskip 2pt}
\ours{} & 0.506 & \textbf{0.012} & \textbf{15.2} & \textbf{27.3} & \textbf{36.4} & \textbf{90.1} & 0.008 \\
\midrule
\rowcolor{gray!20} \multicolumn{8}{c}{\textit{Muse-Glimmer-30B}} \\
\midrule
ReAct & 0.409 & 117.762 & 0.0 & 0.0 & 0.0 & 69.1 & 0.018 \\
Hypothesis Search & 0.394 & 33.911 & 0.0 & 0.0 & 0.0 & 70.4 & 0.017 \\
Popper & 0.370 & 6.320 & 0.0 & 0.0 & 0.0 & 67.9 & 0.018 \\
\hdashline\noalign{\vskip 2pt}
\ours{} & \textbf{0.424} & \textbf{0.376} & \textbf{3.0} & \textbf{6.1} & \textbf{9.1} & \textbf{85.2} & \textbf{0.016} \\
\bottomrule
\end{tabular}%
}
\vspace{-10pt}
\end{table*}

\begin{table*}[t]
\centering
\caption{Experimental results of \ours{} and baseline methods on NewtonBench across three backbone LLMs. Symbolic accuracy (SA~$\uparrow$) is reported for nine combinations of system complexity and equation difficulty, together with overall symbolic accuracy and RMSLE ($\downarrow$).}
\label{tab:newtonbench}
\resizebox{0.9\textwidth}{!}{%
\begin{tabular}{l|ccc|ccc|ccc|cc}
\toprule
\multirow{2}{*}{\textbf{Method}} & \multicolumn{3}{c|}{\cellcolor{nbcol}\textbf{Vanilla Equation}} & \multicolumn{3}{c|}{\cellcolor{nbcol}\textbf{Simple System}} & \multicolumn{3}{c|}{\cellcolor{nbcol}\textbf{Complex System}} & \multicolumn{2}{c}{\cellcolor{nbcol}\textbf{Average}} \\
 & easy & medium & hard & easy & medium & hard & easy & medium & hard & SA (\%) $\uparrow$ & RMSLE $\downarrow$ \\
\midrule
\rowcolor{gray!20} \multicolumn{12}{c}{\textit{Gemini-3.1-Flash-Lite}} \\
\midrule
ReAct & 86.1 & 74.1 & 44.4 & 74.1 & 55.6 & 22.2 & 38.0 & 21.3 & 3.7 & 46.6 & 1.376 \\
Hypothesis Search & 91.7 & 78.7 & 41.7 & 77.8 & 59.3 & 19.4 & 42.6 & 32.4 & 5.6 & 49.9 & 1.167 \\
Popper & \textbf{92.6} & \textbf{85.2} & 48.1 & 88.0 & 61.6 & \textbf{33.3} & 45.4 & 25.0 & 3.7 & 53.7 & 0.985 \\
\hdashline\noalign{\vskip 2pt}
\ours{} & \textbf{92.6} & 83.3 & \textbf{59.3} & \textbf{92.6} & \textbf{79.6} & 31.5 & \textbf{54.6} & \textbf{43.5} & \textbf{11.1} & \textbf{60.9} & \textbf{0.707} \\
\midrule
\rowcolor{gray!20} \multicolumn{12}{c}{\textit{Gemini-3.7-Flash}} \\
\midrule
ReAct & 78.7 & 67.6 & 25.0 & 56.5 & 34.3 & 7.4 & 41.7 & 20.4 & 7.4 & 37.7 & 1.992 \\
Hypothesis Search & 83.3 & 66.7 & 34.3 & 53.7 & 41.7 & 14.8 & 47.2 & 15.7 & 5.6 & 40.3 & 1.884 \\
Popper & 85.2 & 69.4 & 41.7 & 58.3 & 40.7 & 18.5 & 50.0 & 23.1 & 11.1 & 44.2 & 1.832 \\
\hdashline\noalign{\vskip 2pt}
\ours{} & \textbf{90.7} & \textbf{78.7} & \textbf{50.9} & \textbf{63.9} & \textbf{53.7} & \textbf{26.9} & \textbf{57.4} & \textbf{37.0} & \textbf{17.6} & \textbf{53.0} & \textbf{1.501} \\
\midrule
\rowcolor{gray!20} \multicolumn{12}{c}{\textit{Muse-Glimmer-30B}} \\
\midrule
ReAct & 94.4 & \textbf{90.3} & 56.9 & 94.4 & 81.9 & 38.9 & 76.4 & 52.8 & 12.5 & 66.5 & 0.572 \\
Hypothesis Search & \textbf{97.2} & 86.1 & 56.9 & 93.1 & \textbf{88.9} & 36.1 & 80.6 & 48.6 & 15.3 & 67.0 & 0.385 \\
Popper & 95.8 & 88.9 & 66.7 & 93.1 & 79.2 & 43.1 & 68.1 & 48.6 & 15.3 & 66.5 & 0.408 \\
\hdashline\noalign{\vskip 2pt}
\ours{} & 94.4 & \textbf{90.3} & \textbf{70.8} & \textbf{97.2} & 83.3 & \textbf{51.4} & \textbf{83.3} & \textbf{61.1} & \textbf{16.7} & \textbf{72.1} & \textbf{0.176} \\
\bottomrule
\end{tabular}%
}
\end{table*}

\section{Analysis}\label{sec:analysis}
Beyond the accuracy of the submitted laws, we further analyze how \ours{} achieves these gains. We first study the effect of the number of candidate hypotheses and examine its recorded inference usage. We then compare the error rates of ReAct and \ours{} to identify which failures become less frequent. Finally, we conduct ablation studies on the components of \ours{}.

\subsection{Effect of the Number of Candidate Hypotheses}\label{sec:k}
\begin{figure}[t]
\centering
\includegraphics[width=1.0\columnwidth]{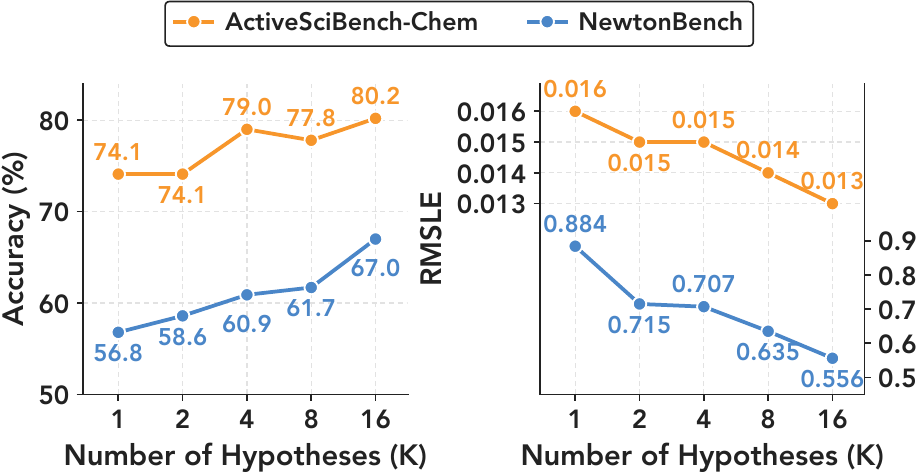}
\caption{Effect of the number of candidate hypotheses $K$ on ActiveSciBench-Chem and NewtonBench with Gemini-3.1-Flash-Lite. (a) accuracy (exact accuracy for ActiveSciBench-Chem and symbolic accuracy for NewtonBench); (b) RMSLE. The main experiments use $K=4$.}
\label{fig:kscan}
\vspace{-10pt}
\end{figure}

The number of candidates $K$ in Eq.~\ref{eq:sample} controls how broadly \ours{} explores at each proposal step, at a sampling cost that grows with $K$. We vary $K$ from 1 to 16. As shown in Figure~\ref{fig:kscan}, the exact accuracy on ActiveSciBench-Chem rises from 74.1\% for $K \le 2$ to 79.0\% at $K=4$ and changes little beyond this point, and NewtonBench follows the same trend. These results suggest that \ours{} can identify promising directions from a broader set of candidate hypotheses.

\subsection{Cost-Effectiveness Analysis}\label{sec:cost}
\begin{figure}[t]
\centering
\includegraphics[width=0.9\columnwidth]{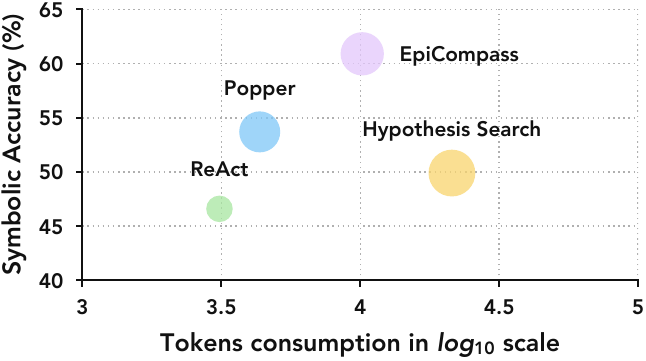}
\caption{Preliminary accuracy--usage comparison on NewtonBench with Gemini-3.1-Flash-Lite. Dot size denotes recorded LLM calls per task; the axes show recorded tokens per task on a $\log_{10}$ scale and symbolic accuracy (\%). Accounting limitations are described in the text.}
\label{fig:cost}
\vspace{-10pt}
\end{figure}

We compare the performance and token consumption of \ours{} with the baselines on NewtonBench using Gemini-3.1-Flash-Lite as the backbone. As shown in Figure~\ref{fig:cost}, \ours{} achieves higher symbolic accuracy than Hypothesis Search (60.9\% vs.\ 49.9\%) while using fewer recorded tokens per task (10.2K vs.\ 21.4K). These results suggest that explicitly tracking epistemic states can improve discovery performance without relying solely on increased token consumption.

\subsection{Error Analysis}\label{sec:error}
\begin{figure}[t]
\centering

\includegraphics[width=1.0\columnwidth]{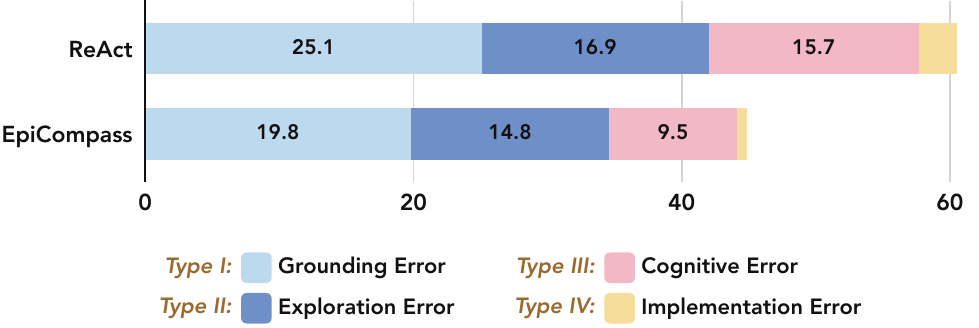}
\caption{Error analysis of four failure categories for ReAct and \ours{} (y-axis), measured as percentages of all analyzed trajectories (x-axis). Results are obtained on NewtonBench with Gemini-3.1-Flash-Lite.}
\label{fig:error}
\vspace{-10pt}
\end{figure}

To investigate how epistemic state tracking affects discovery behavior, we analyze all the trajectories on NewtonBench with Gemini-3.1-Flash-Lite and assign the four error categories defined in \S\ref{sec:failure-modes}. Here, each category percentage uses all 243 analyzed trajectories as its denominator, including successful trajectories; these are error rates rather than the normalized error composition in Figure~\ref{fig:failure-modes}.
As shown in Figure~\ref{fig:error}, \ours{} reduces grounding errors from 25.1\% to 19.8\%, exploration errors from 16.9\% to 14.8\%, and cognitive errors from 15.7\% to 9.5\%. The largest absolute reduction occurs in cognitive errors, consistent with the role of explicit belief updates in helping agents reconsider hypotheses when confronted with conflicting evidence. The reduction in grounding errors suggests that maintaining hypothesis--evidence relationships helps agents keep their claims consistent with experimental observations. Exploration errors decrease more modestly, suggesting that identifying unresolved gaps may be insufficient to produce more informative hypotheses. Overall, the improvements are more pronounced in grounding claims and revising beliefs than in broadening exploration, highlighting the continuing challenge of turning an explicit exploration state into productive scientific inquiry.

\subsection{Ablation Study}\label{sec:ablation}
\begin{table*}[t]
\centering
\caption{Cumulative ablation study of \ours{} components with Gemini-3.1-Flash-Lite. Each row further removes one component from the row above it. The best results are highlighted in \textbf{bold}. The first three NewtonBench columns report symbolic accuracy (\%) averaged within each system complexity.}
\label{tab:ablation}
\resizebox{\textwidth}{!}{%
\begin{tabular}{l|cccc|cc|cc}
\toprule
\multirow{2}{*}{\textbf{Components}} & \multicolumn{4}{c|}{\cellcolor{nbcol}\textbf{NewtonBench}} & \multicolumn{2}{c|}{\cellcolor{dpcol}\textbf{DiscoverPhysics}} & \multicolumn{2}{c}{\cellcolor{sgcol}\textbf{ActiveSciBench-Chem}} \\
 & Vanilla & Simple & Complex & SA (\%) $\uparrow$ & nMSE $\downarrow$ & Expl. $\uparrow$ & Acc. (\%) $\uparrow$ & RMSLE $\downarrow$ \\
\midrule
\ours{} (Full) & 78.4 & \textbf{67.9} & \textbf{36.4} & \textbf{60.9} & 1.42 & \textbf{0.379} & \textbf{79.0} & \textbf{0.015} \\
\quad w/o Belief Update & \textbf{82.4} & 63.0 & 27.8 & 57.7 & \textbf{1.39} & 0.370 & 67.9 & 0.021 \\
\quad w/o Epistemically Grounded Self-Verification & 74.1 & 59.3 & 30.6 & 54.6 & 4.32 & 0.355 & 64.2 & 0.021 \\
\quad w/o Epistemic-Gain Hypothesis Exploration & 70.4 & 50.9 & 22.2 & 47.8 & 25.46 & 0.300 & 64.2 & 0.023 \\
\bottomrule
\end{tabular}%
}
\vspace{-10pt}
\end{table*}

We conduct cumulative ablation studies to analyze the contribution of each component in \ours{}. As shown in Table~\ref{tab:ablation}, removing belief updates reduces symbolic accuracy on NewtonBench from 60.9\% to 57.7\% and exact accuracy on ActiveSciBench-Chem from 79.0\% to 67.9\%, highlighting the importance of reassessing competing hypotheses as evidence accumulates. Further removing epistemically grounded self-verification lowers accuracy on both benchmarks and increases normalized trajectory error on DiscoverPhysics from 1.39 to 4.32. This supports the value of checking the proposed answer against the complete epistemic state before submission. Finally, removing candidate generation and selection based on epistemic gain reduces symbolic accuracy on NewtonBench to 47.8\% and increases trajectory error on DiscoverPhysics to 25.46, underscoring the importance of exploring candidates that combine evidential support with promising alternatives. Together, these results support the complementary roles of updating current beliefs, exploring alternative explanations, and verifying the final conclusion. 

\section{Related Work}

\textbf{Autonomous Scientific Discovery.}
Language agents increasingly conduct research end to end. Goal-driven systems carry out the research cycle from a specified question~\citep{yamada2025ai, jansen2025codescientist, schmidgall2025agent, baek2025researchagent, zhao2026scienceflow} or validate a given hypothesis through sequential falsification~\citep{huang2025automated, liu2025hypobench}. A complementary line explores \emph{open-ended} discovery, granting agents greater autonomy in formulating hypotheses and research questions. This includes environments supporting the full discovery cycle~\citep{jansen2024discoveryworld}, benchmarks combining hypothesis proposal with executable verification~\citep{majumder2025discoverybench, duan2026measuring}, agents that generate and refine hypotheses across large search spaces~\citep{majumder2024data, wang2024hypothesis, gupta2026accelerating}, and methods that guide experimentation through changes in the agent's beliefs~\citep{agarwal2026autodiscovery, agarwal2026evidence}. Studies on real research tasks nevertheless reveal overstated conclusions, fixation on familiar assumptions, and shifts in direction without sufficient follow-up~\citep{trehan2026why, kirgis2026can, fei2026agents, du2026station}. Position papers further argue that current agents prioritize solving predefined problems over discovery and lack a persistent, revisable representation of their knowledge~\citep{bisht2026agentic}. \ours{} addresses this gap through a plug-and-play scaffold that explicitly tracks epistemic states to guide hypothesis revision and exploration, evaluated in interactive discovery settings. \nocite{yuan2026flowbank,yuan2026meta}

\textbf{Belief and Epistemic States in Agents.}
Prior work examines how LLMs express uncertainty~\citep{lin2022teaching,tian2023just,xiong2024can}, estimate confidence~\citep{kadavath2022language,kuhn2023semantic}, and revise beliefs as evidence arrives~\citep{wilie2024belief}. However, their judgments remain sensitive to user pressure~\citep{sharma2024towards,perez2023discovering}, their stated reasoning may be unfaithful~\citep{turpin2023language}, and their probability judgments can violate coherence~\citep{zhu2024incoherent,binz2023using}. These limitations motivate explicit representations of agents' beliefs and supporting evidence. For long-horizon tasks, PoS maintains belief states with consistency checks~\citep{luo2026beyond}, while Agent-BRACE learns a separate belief model with verbalized certainty~\citep{singh2026brace}. In scientific discovery, HEP maintains a hypothesis registry with probabilities and lifecycle states~\citep{takahara2026auditable}, BayesEvolve tracks uncertainty over hypothesis quality~\citep{wu2026bayesevolve}, StatefulDiscovery grounds claims in evidence~\citep{chen2026statefuldiscovery}, and Arbor refines hypothesis trees through insight back-propagation~\citep{jin2026arbor}. Nevertheless, discovery traces reveal neglected evidence and limited revision after refutation~\citep{rios2026ai}. Our work connects belief tracking with an explicit \emph{exploration state}: unresolved conflicts and explanatory gaps guide hypothesis generation and qualitative epistemic-gain assessment, while final verification checks the conclusion against the accumulated epistemic tree.

\section{Conclusion}
We introduce Epistemic Compass (\ours), a plug-and-play scaffold that maintains an explicit epistemic state for automated scientific discovery. Through an epistemic tree, \ours{} tracks hypotheses, evidence, beliefs, and revisions, using unresolved conflicts and unexplored directions to guide hypothesis generation and final verification. Across three scientific discovery benchmarks and three LLM backbones, \ours{} consistently improves discovery performance under matched experimental budgets. Further analyses show fewer unsupported claims and failures to revise beliefs, while cumulative ablations support the contributions of belief updates, hypothesis selection, and self-verification.

\noindent\textbf{Discussion.}
Scientific discovery depends on how accumulated evidence changes both what an agent believes and what it investigates next. An explicit epistemic state makes these changes inspectable and provides a basis for guiding exploration beyond existing hypotheses. Future work can study how agents learn from their evolving epistemic states to pursue more productive research directions.

\section*{Impact Statement}
This work aims to make LLM-based scientific discovery agents more reliable and more transparent. Externalizing the epistemic state lets human experts audit how an agent reached a conclusion, which hypotheses it overturned, and which questions it left open, reducing the risk of confidently wrong automated findings. The method does not alter model parameters and inherits the limitations and biases of the underlying LLM; conclusions produced by discovery agents should remain subject to expert validation before use in high-stakes scientific or engineering decisions.

\bibliography{example_paper}
\bibliographystyle{icml2027}

\newpage
\appendix
\onecolumn

\section*{Appendix Outline}
This appendix provides the discovery algorithm and implementation details, benchmark settings and evaluation metrics, supplementary analysis of agent deliberation, and the prompts used by the framework.

\newcommand{\appentry}[3]{%
  \noindent\hspace*{#1}\hyperref[#2]{\ref*{#2}\quad #3}\nobreak\dotfill\nobreak\hyperref[#2]{\pageref*{#2}}\par}
\begingroup
\setlength{\parskip}{0.45em}
\appentry{0pt}{app:algorithm}{\textbf{Algorithmic and Implementation Details}}
\appentry{1.5em}{app:procedure}{Discovery Procedure}
\appentry{1.5em}{app:feedback}{Epistemic Feedback Settings}
\appentry{0pt}{app:benchmarks}{\textbf{Benchmarks and Evaluation Metrics}}
\appentry{1.5em}{app:physics}{DiscoverPhysics}
\appentry{1.5em}{app:chem}{ActiveSciBench-Chem}
\appentry{1.5em}{app:newton}{NewtonBench}
\appentry{0pt}{app:analysis}{\textbf{Supplementary Analysis}}
\appentry{1.5em}{app:steps}{Deliberation before Submission}
\appentry{0pt}{app:prompts}{\textbf{Prompts}}
\endgroup
\clearpage

\section{Algorithmic and Implementation Details}\label{app:algorithm}
\subsection{Discovery Procedure}\label{app:procedure}
Algorithm~\ref{alg:ours} summarizes the interaction loop described in Section~\ref{sec:method}. Beliefs are defined once the hypothesis set is nonempty. The routine state summary conveys both the evidential status of explored hypotheses and the unresolved directions represented by $U_t$.

\begin{algorithm}[H]
\caption{Scientific Discovery with \ours{}}
\label{alg:ours}
\begin{algorithmic}[1]
\Require agent policy $\pi$, candidates $K$, experiment budget $B$
\State $\mathcal{D}_0,H_0,U_0,V_0,\mathcal{T}_0\gets\emptyset$; $b_0\gets\mathrm{undefined}$; $s_0\gets\varnothing$
\State $\mathcal{E}_0\gets(\mathcal{D}_0,H_0,b_0,U_0)$; initialize $\tau_0$ and remaining budget
\For{$t=1,2,\dots$ while the benchmark permits another interaction}
  \State Copy $\mathcal{D}_{t-1},H_{t-1},b_{t-1},U_{t-1},V_{t-1},\mathcal{T}_{t-1}$ to their step-$t$ counterparts
  \State $s_t\gets\varnothing$; $a_t\sim\pi(\cdot\mid\tau_{t-1},s_{t-1})$
  \If{$a_t$ is a hypothesis proposal}
    \State Sample $K$ candidates and replace the proposal with selected $h_t$ \Comment{Eqs.~\ref{eq:sample}, \ref{eq:select}}
    \State $H_t\gets H_t\cup\{h_t\}$; attach $h_t$ to $\mathcal{T}_t$
  \ElsIf{$a_t$ is an experimental testing action}
    \State Execute within budget $B$; add observed experiment--outcome pairs to $\mathcal{D}_t$
  \ElsIf{$a_t$ commits $\hat h$}
    \If{this is the first submission}
      \State $s_t\gets\textsc{Render}(\mathcal{T}_t)$ in full; request verification \Comment{\S\ref{sec:verify}}
    \Else
      \State \Return $\hat h$
    \EndIf
  \EndIf
  \If{new evidence or a hypothesis was added, and $H_t\ne\emptyset$}
    \State Update $V_t$; revise $b_t$ and exploration directions $U_t$; refresh $\mathcal{T}_t$ and present $s_t$ subject to the routine feedback limit \Comment{\S\ref{sec:belief}}
  \EndIf
  \State $\mathcal{E}_t\gets(\mathcal{D}_t,H_t,b_t,U_t)$
  \State $\tau_t\gets\tau_{t-1}$ appended with the executed action, its response, and any delivered feedback
\EndFor
\State \Return the final hypothesis under the benchmark's termination rules
\end{algorithmic}
\end{algorithm}

\subsection{Epistemic Feedback Settings}\label{app:feedback}
Each routine epistemic state summary includes up to five hypotheses and two conflicting experiments per hypothesis; final verification presents the complete tree. Candidate selection and belief updates use the same backbone as the discovery agent. We use $K=4$ candidate hypotheses and cap routine epistemic feedback at eight presentations per episode, as described in Section~\ref{sec:impl}.

\section{Benchmarks and Evaluation Metrics}\label{app:benchmarks}

We describe the benchmark settings and evaluation metrics below. All metrics are computed on the final submitted hypotheses using the official evaluation scripts with the evaluation models specified below.

\subsection{DiscoverPhysics}\label{app:physics}
\noindent\textbf{Task and Settings.}
DiscoverPhysics~\citep{wiemann2026discoverphysics} evaluates the discovery of physical laws in simulated worlds that deviate from canonical physics. The agent specifies particle initial conditions, observes noisy trajectories from an $N$-body simulator, and submits an executable force law with a natural-language explanation. We evaluate on 11 public worlds with a budget of 16 experiment rounds and 5\% observation noise. Each world is run with three noise seeds, yielding 33 scheduled runs per method and backbone.

\noindent\textbf{Metrics.}
\textit{Normalized trajectory MSE (nMSE).} Free parameters of the submitted law are fitted to the observed data, and predictions are evaluated on held-out, noise-free trajectories. Position MSE is normalized by the variance of the ground-truth trajectories. We report the geometric mean across runs; non-executable submissions receive a large constant error.

\textit{Explanation score (Expl.).} Gemini-4-Argon evaluates explanations against expert-written rubrics on a scale of 0--10, rescaled to $[0,1]$.

\textit{Pass@$k$.} A run succeeds if its nMSE is below 0.1 and its explanation score is at least 0.9. We report the expected percentage of worlds with at least one successful run among $k$ runs sampled without replacement, for $k \in \{1,2,3\}$.

\subsection{ActiveSciBench-Chem}\label{app:chem}
\noindent\textbf{Task and Settings.}
ActiveSciBench-Chem~\citep{kabra2026llm} evaluates the discovery of enzyme reaction rate laws from observations
$r_{\mathrm{obs}}=f(C_A,C_I,C_B,C_P,\mathrm{Enz},T,\mathrm{pH})$,
where the inputs describe reactant concentrations, enzyme concentration, temperature, and pH. The agent selects experimental conditions, observes initial reaction rates, and submits the underlying concentration-dependent rate law; temperature and pH contribute background effects as described below. We use 27 mechanisms at three difficulty levels with law version v0, yielding 81 tasks without observation noise. Agents are given the possible mechanism families but not the active mechanism.

All methods use the same agent loop with at most 12 rounds, 10 experiments per round, and 60 experiments in total. Observed rates include deterministic temperature and pH background effects of approximately $\pm5\%$, which are excluded from the target law. Accordingly, \ours{} uses a 6\% relative-error threshold when labeling supporting evidence. Gemini-3.1-Flash-Lite and Muse-Glimmer-30B use the official prompt; Gemini-3.7-Flash receives an additional clarification about these background effects and the evaluation target, identical across methods.

\noindent\textbf{Metrics.}
The submitted law is evaluated against the noise-free target on 1,000 test points. \textit{Exact accuracy (Acc.)} is the percentage of tasks with RMSLE below 0.05. We also report the median \textit{RMSLE} across tasks, treating unevaluable submissions as the worst-performing cases.

\subsection{NewtonBench}\label{app:newton}
\noindent\textbf{Task and Settings.}
NewtonBench~\citep{zheng2026newtonbench} evaluates the discovery of counterfactually modified physical laws across 12 domains. Each target law appears either as a standalone equation or within a simple or complex system containing additional known equations. Each domain includes 27 configurations spanning three equation difficulties, three law versions, and three system complexities. We evaluate each configuration without observation noise, using three trials for each Gemini backbone (972 trials) and two for Muse-Glimmer-30B (648 trials), and average the metrics across configurations and runs.

\noindent\textbf{Metrics.}
\textit{Symbolic accuracy (SA).} Gemini-4-Argon judges whether the submitted law is mathematically equivalent to the target law, disregarding the values of physical constants.

\textit{RMSLE.} Predictive error is evaluated on 5,000 inputs sampled from domain-specific distributions:
\begin{equation*}
\mathrm{RMSLE}
=
\sqrt{\frac{1}{N}\sum_{i=1}^{N}
\left[\log(\hat{y}_i+1)-\log(y_i+1)\right]^2}.
\end{equation*}
Outliers are filtered using the modified $z$-score, following the official implementation.

\section{Additional Analysis}\label{app:analysis}

\subsection{Round of Iterations}\label{app:steps}

\begin{wrapfigure}{r}{0.4\textwidth}
\vspace{-59pt}
\centering
\includegraphics[width=\linewidth]{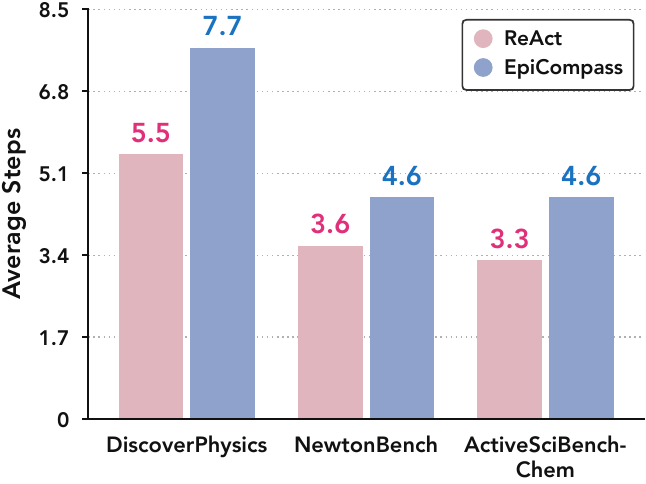}
\caption{Average interaction rounds before final submission with Gemini-3.1-Flash-Lite across three benchmarks.}
\label{fig:steps}
\vspace{-20pt}
\end{wrapfigure}

We examine the number of interaction rounds before final submission to assess how \ours{} affects the investigation process. As shown in Figure~\ref{fig:steps}, \ours{} takes more rounds than ReAct on all three benchmarks with Gemini-3.1-Flash-Lite, increasing the average from 5.5 to 7.7 on DiscoverPhysics. The recorded results show the same trend for the other two backbones. On NewtonBench and ActiveSciBench-Chem, ReAct averages at most 1.2 hypothesis rounds across backbones, suggesting limited hypothesis revision before submission. With Gemini-3.1-Flash-Lite, \ours{} increases these averages to 1.5 and 1.9, respectively, while its additional rounds on DiscoverPhysics are primarily devoted to experiments. Together with the accuracy improvements in Tables~\ref{tab:main} and~\ref{tab:newtonbench}, these results suggest that epistemic state tracking encourages further investigation and revision within the fixed experiment budget, rather than early commitment to an initial hypothesis. \nocite{yuan2026flowbank, yuan2026meta, yu2026weak, wang2026mitigating}

\clearpage

\section{Prompts}\label{app:prompts}

\newtcblisting{promptbox}[1]{%
  breakable,
  listing only,
  colback=gray!4,
  colframe=gray!55,
  boxrule=0.5pt,
  arc=1pt,
  left=6pt,
  right=6pt,
  top=4pt,
  bottom=4pt,
  title={#1},
  fonttitle=\bfseries\small,
  coltitle=black,
  colbacktitle=gray!15,
  listing options={%
    basicstyle=\ttfamily\scriptsize,
    breaklines=true,
    breakatwhitespace=false,
    columns=fullflexible,
    keepspaces=true,
    extendedchars=true,
    upquote=true
  }
}

We provide the prompts and the Open Directions feedback template used by \ours{}. All prompt template fields are populated dynamically from the agent's interaction trajectory and maintained epistemic state. Candidate proposal evaluation, belief revision, and epistemic summary formatting use the same foundational backbone as the discovery agent. Within each benchmark--backbone evaluation, all comparative baselines share the uniform base prompt structure, as detailed in Appendix~\ref{app:benchmarks}.

\subsection{Prompts of \ours{}}

\noindent\textbf{Candidate Generation and Selection.}
At each hypothesis proposal step, \ours{} samples $K$ candidate hypotheses conditioned on the interaction history and epistemic feedback. Each candidate is evaluated against accumulated experimental observations and contextualized within the existing epistemic hierarchy. The selection policy model then selects one candidate to maximize anticipated epistemic gain, balancing evidential support against its potential to resolve outstanding explanatory gaps.

\begin{promptbox}{Prompt: Hypothesis Selection by Epistemic Gain}
A scientist is investigating an unknown system. Below is the current state of their investigation -- the hypotheses they have written so far and how their own experiments bear on each -- followed by {k} candidate new hypotheses they could propose at this step, each with how the same experiments bear on it and how it relates to the existing hypotheses. All facts are computed from the scientist's own experiments.

Choose the ONE candidate that most opens a promising new direction while being compatible with the evidence so far:
1. Prefer a candidate that can account for what the existing hypotheses cannot, without contradicting what the evidence already supports.
2. A restatement of an existing hypothesis, or a direction the experiments already contradict, is less valuable.
3. Do not judge by which candidate you believe is the true law; judge strictly by the evidence and the exploratory direction.

CURRENT STATE:
{state}

CANDIDATES:
{cands}

Return JSON only:
{"choice": <candidate number>, "why": "<one sentence>"}
\end{promptbox}

\noindent\textbf{Belief Update.}
Upon acquiring new experimental evidence or retaining a proposed hypothesis, \ours{} updates the relative credence distribution over all active hypotheses, provided the hypothesis set is non-empty. The policy model conditions on prediction errors, empirical consistency, and model complexity to assign relative beliefs summing to 1.0.

\begin{promptbox}{Prompt: Belief Update over Retained Hypotheses}
A scientist is studying an unknown system. Below are the candidate hypotheses they wrote and, as evidence, how each of the scientist's own experiments bears on each hypothesis: its prediction error ({unit}) and whether the experiment supports it, conflicts with it, or gives no evidence either way ("fails" = no finite prediction).

Based on this evidence, state your belief that each hypothesis is the correct one among these candidates: a number in [0, 1] per hypothesis, summing to 1.0.

Weigh how well each fits all of the experiments, penalize any experiment it fails or misses badly, and note that a hypothesis with more free parameters fits more easily (prefer simpler hypotheses when fit is comparable).

HYPOTHESES:
{hyps}

EVIDENCE (rows = experiments, columns = hypotheses):
{table}

Return JSON only:
{"beliefs": {{ids}}}
\end{promptbox}

\noindent\textbf{Epistemic Feedback.}
The policy model compiles the active epistemic state into a structured summary returned to the discovery agent. The formatting pipeline strictly restates verified observations without introducing speculative hypotheses or prescriptive experimental interventions.

\begin{promptbox}{Prompt: Epistemic State Summarization}
You maintain a scientist's epistemic-state card. Below are FACTS computed from the scientist's own experiments, hypotheses, and statements, grouped by section.

Write the card as JSON with exactly these keys:
- "Hypotheses"
- "Conflicts"
- "Your claims"
- "Still undecided"
- "Changed"
(each mapped to a list of short bullet strings).

Rules:
1. Restate only what the FACTS say, keeping every hypothesis ID, number, experiment input, and quote exactly as given.
2. You may merge, shorten, and order the items by importance.
3. Leave a key's list empty ([]) if it has no facts.
4. Do NOT say which hypothesis is right, do NOT suggest a law, a mathematical form, a parameter value, or an experiment that is not in the FACTS, and add no advice.

FACTS:
{facts}

Return JSON only.
\end{promptbox}

\noindent\textbf{Exploration Update and Open Directions.}
The Open Directions section makes the exploration state explicit through experiments unexplained by any retained hypothesis, inputs that have not been varied or have been tested at only two values, and hypothesis pairs that existing experiments do not distinguish. The template lists at most four unexplained experiments and six inputs in each coverage category. Here, \texttt{\{k\}} denotes the number of unexplained experiments.

\begin{promptbox}{Feedback Template: Open Directions}
[Epistemic state - computed only from your own experiments, your own models/laws, and your own statements; it does not say which hypothesis is right]

## Open directions
- {k} of your {n} experiments are explained by none of your hypotheses: {experiment}; {experiment}; ...
- Inputs you have not varied yet: {input} (always {value}), ...
- Inputs you have varied at only two values: {input}, {input}, ...
- None of your experiments separates {H_a} and {H_b}: they agree on every experiment so far.
\end{promptbox}

\noindent\textbf{Epistemically Grounded Self-Verification.}
At the first submission attempt, \ours{} presents the complete epistemic dependency tree together with the message below. The agent may revise its hypothesis, conduct further experiments within the remaining budget, or submit it unchanged. The subsequent submission is treated as final.

\begin{promptbox}{System Notice: Self-Verification at Initial Commit}
Before this submission is scored, you may revise it, investigate further if you have turns left, or submit it unchanged. Your next submission is final. You have {n} experiment turn(s) left.
\end{promptbox}

\end{document}